\documentclass{article}
\usepackage[T1]{fontenc}

\usepackage{microtype}
\usepackage{graphicx}
\usepackage{subfigure}
\usepackage{booktabs} 

\usepackage{hyperref}

\usepackage[accepted]{icml2025}

\usepackage{amsmath}
\usepackage{amssymb}
\usepackage{mathtools}
\usepackage{amsthm}

\usepackage{makecell}
\usepackage{tikz}
\usepackage{float}
\usepackage{booktabs}
\usepackage{graphicx}
\usepackage{multirow}
\usepackage{pgfplots}
\pgfplotsset{compat=1.17}
\usepgfplotslibrary{groupplots}

\usepackage[capitalize,noabbrev]{cleveref}

\theoremstyle{plain}

\theoremstyle{definition}

\theoremstyle{remark}

\usepackage[textsize=tiny]{todonotes}

\icmltitlerunning{Scaling Natural-Language Graph-Based Test Time Compute for Automated Theorem Proving}

\begin{document}

\twocolumn[
\icmltitle{Scaling Natural-Language Graph-Based Test Time Compute for Automated Theorem Proving}



\icmlsetsymbol{equal}{*}

\begin{icmlauthorlist}
\icmlauthor{Vincent Li}{equal,bu}
\icmlauthor{Tim Knappe}{equal,provadis}
\icmlauthor{Yule Fu}{duke}
\icmlauthor{Kevin Zhu}{algo}
\icmlauthor{Kevin Han}{algo}
\end{icmlauthorlist}

\icmlaffiliation{bu}{Boston University}
\icmlaffiliation{provadis}{Provadis School of International Management \& Technology}
\icmlaffiliation{duke}{Duke University}
\icmlaffiliation{algo}{Algoverse AI Research}

\icmlcorrespondingauthor{Vincent Li}{vinli@bu.edu}

\icmlkeywords{Machine Learning, ICML, Large Language Models, mathematical reasoning, knowledge graphs, proof formalization, natural language understanding, theorem proving}

\vskip 0.3in
]



\printAffiliationsAndNotice{
The second AI for MATH Workshop at the 42nd International Conference on Machine Learning, Vancouver, Canada
\icmlEqualContribution
} 

\begin{abstract}
Large language models have demonstrated remarkable capabilities in natural language processing tasks requiring multi-step logical reasoning capabilities, such as automated theorem proving. However, challenges persist within theorem proving, such as the identification of key mathematical concepts, understanding their interrelationships, and formalizing proofs correctly within natural language. We present KG-prover, a novel framework that leverages knowledge graphs mined from reputable mathematical texts to augment general-purpose LLMs to construct and formalize mathematical proofs. We also study the effects of scaling graph-based, test-time compute using KG-Prover, demonstrating significant performance improvements over baselines across multiple datasets. General-purpose LLMs improve up to 21\% on miniF2F-test when combined with KG-Prover, with consistent improvements ranging from 2-11\% on the ProofNet, miniF2F-test, and MUSTARD datasets. Furthermore, KG-Prover with o4-mini achieves 50\% on pass miniF2F-test. This work provides a promising approach for augmenting natural language proof reasoning with knowledge graphs without the need for additional finetuning.

\end{abstract}

\section{Introduction}

The advent of Large Language Models has revolutionized natural language processing, enabling machines to perform complex reasoning tasks using transformer models \citep{vaswani2023attentionneed, peters-etal-2018-deep, brown2020languagemodelsfewshotlearners, srivastava2023imitationgamequantifyingextrapolating}. Transformer-based models have shown promise in mathematical problem-solving, which inherently requires multi-step logical inference and a precise understanding of abstract concepts \citep{Robinson2001, Guo2025}. Despite these advancements, significant challenges remain in automating the identification of mathematical concepts, understanding their interrelations, and formalizing proofs within a mathematical framework \citep{hendrycks2021measuringmathematicalproblemsolving}.
Work by \cite{Polu2020} introduced training language models to generate proofs in formal languages and use such models to address the generation of original mathematical terms -- leading to the introduction of the GPT-f proof assistant for the Metamath formalization language.
Systems such as InternLM2.5-StepProver and DeepSeek-Prover-V2 directly generate proof candidates in the Lean language, achieving state-of-the-art performance in a wide variety of theorem proving benchmarks \cite{wu2024internlm25stepproveradvancingautomatedtheorem, ren2025deepseekproverv2advancingformalmathematical}.

Recent advances in AI-driven mathematics have targeted the integration of neurosymbolic architectures with formal verification frameworks. Systems such as DeepMath and HOList employ MCTS guided by graph neural networks to prune combinatorial proof spaces \cite{Bansal2019, alemi2017deepmathdeepsequence}. These frameworks combine self-play reinforcement learning with backward-chaining, enabling exploration of lemma sequences in interactive theorem provers.

A parallel line of research explores the use of natural language as an intermediate representation for guiding formal reasoning. Notably, \citet{jiang2023draftsketchproveguiding} introduced a draft-sketch-prove pipeline, in which informal proof sketches are first generated in natural language and then incrementally translated into formal code. This enables the model to exploit the flexibility of natural reasoning, though at the cost of potential errors and ambiguity during the translation into a formal theorem proving language such as Lean. TheoremLlama attempts to bridge the gap between natural language (NL) reasoning and formal language (FL) proofs using an NL-FL aligned dataset for training while still integrating NL text in the proof \cite{wang2024theoremllamatransforminggeneralpurposellms}.


In this work, we introduce KG-Prover, a novel automated theorem-proving framework allowing a general purpose LLM to semantically retrieve and traverse a knowledge graph derived from ProofWiki.

KG-Prover begins with selecting a starting node via semantic embedding lookup, followed by iteratively and selectively expanding the traversal of covered nodes by returning the top-k most similar nodes. In the beam search method, we implement LLMs as proof judges that return the most promising candidates \cite{zhu2025judgelmfinetunedlargelanguage}. The general-purpose LLM then generates an informal proof which is then autoformalized and verified using Lean \cite{Moura2021}. We show that effective test time compute scaling is achievable by simply modulating the graph traversal depth.

Unlike previous approaches, our framework does not rely on large amounts of formal training data or intensive expert iteration. Instead, we operate with no specialized training, leveraging the built-in natural language reasoning abilities of general-purpose LLMs to synthesize graph-retrieved information before outputting an informal proof.

Our contributions are as follows:
\begin{itemize}
    \item Develop KG-Prover, an automated theorem proving framework relying on natural language informal proof generation combined with an iterative refinement-based knowledge graph traversal and an LLM as a judge.
    \item We build a knowledge graph using ProofWiki of over 60,000 nodes and 300,000 edges that represent mathematical concepts and their interrelations, modeling complex relationships with mathematically similar subjects.
    \item We introduce an iterative refinement system based on a heuristic evaluation by a model judge and beam search for further revisions, improving performance by up to 26.4\% over baseline and 21.8\% over the non-scaling KG-Prover.
\end{itemize}

\section{Related Work}
\textbf{Learning-Based Formal Provers} Recent advancements in theorem proving have increasingly focused on integrating structured knowledge with LLMs. Notably, DeepSeek-Prover-V1.5 \cite{Xin2024DeepSeekProverV15HP} combines reinforcement learning from proof assistant feedback (RLPAF) with Monte-Carlo tree search. The model, pre-trained on formal mathematical languages like Lean 4, achieves state-of-the-art results on miniF2F-test and ProofNet. It does so by dynamically exploring diverse proof paths through intrinsic-reward-driven search. This builds on earlier work such as LeanDojo \cite{Yang2023LeanDojoTheoremProving}, which introduced ReProver, an LLM-based prover enhanced with retrieval capabilities to efficiently select theorem premises. Similarly, HyperTree Proof Search \cite{Lample2022HyperTree} demonstrated that structured search algorithms could enhance proof generation in formal systems like Metamath.
Furthermore, \citet{wu2022autoformalizationlargelanguagemodels} showed that LLMs can effectively translate informal mathematical statements into formal logic, targeting Isabelle/HOL, proving that the resulting autoformalized specifications are sufficiently accurate to improve downstream formal provers trained on them.

\textbf{Sampling and Compute Strategies} Additionally, \cite{Snell2024ScalingLLMTestTime} proposes a "compute-optimal" strategy that dynamically adjusts resources based on task difficulty. This approach achieves efficiency gains over traditional sampling and allows smaller models to outperform larger counterparts in FLOPs-matched evaluations. The strategy is broadly applicable in various complex reasoning domains, including automated theorem proving.

\textbf{Feedback Mechanisms and Self-Improving Agents} In improving feedback mechanisms, STP \cite{Lee2025STPSelfplayLLMTheorem} uses self-play between conjecturer and prover agents, while Formal Theorem Proving by Hierarchical Decomposition \cite{Sun2024FormalTheoremProving} rewards lemma decomposition via reinforcement learning. Finally, the MUSTARD project \citep{huang2024mustardmasteringuniformsynthesis} used an iterative approach where the LLM generates a problem, constructs an informal proof, converts it into Lean \citep{Moura2015} format, and verifies the proof with a Lean interpreter. MUSTARD operates in three stages: sampling concepts, using generative models to create problems and solutions, and employing proof assistants to validate these solutions.
\citet{jiang2023draftsketchproveguiding} proposed a three-phase framework that first drafts an informal proof in natural language, then sketches a rough tactic script in Lean, and finally invokes a formal prover to complete the remaining subgoals. This approach illustrates how guidance can yield strong formal results, supporting our use of informal proof generation as a first-class component.

\textbf{Graph LMs and Retrieval Mechanisms} Graph-based retrieval-augmented generation techniques have also received growing attention for their ability to leverage structured relationships to enhance downstream tasks such as question answering and formal proof search. For instance, GraphRetriever builds a passage graph whose edges are derived from an external knowledge base and uses it to retrieve relevant passages for open-domain question answering, outperforming text-only retrieval baselines \citep{Wang2022}. Similarly, QA-GNN jointly reasons over a retrieved question-relevant knowledge subgraph and language-model representations through a graph neural network, enabling more interpretable and accurate reasoning within language model generation \citep{Verma2023}. Beyond question answering, GRAG retrieves textual subgraphs and integrates the joint textual and topological information into LLMs to enhance generation \citep{Zhao2021}. Collectively, these works underscore the capabilities of knowledge graphs with LLMs for reasoning tasks, providing more effective retrieval.

\textbf{Lean Provers}
Recent works in direct Lean proving have shown promising advances in consistently formalizing correct and rigorous mathematical proofs. By training and finetuning LMs such as InternLM2.5-StepProver and DeepSeek-Prover-V2 to generate directly in Lean's formal language, these systems demonstrate state-of-the-art performance in autoformalization tasks \cite{wu2024internlm25stepproveradvancingautomatedtheorem, ren2025deepseekproverv2advancingformalmathematical}.
InternLM 2.5–StepProver applies expert iteration entirely within Lean, using curriculum learning and self-generated proofs to continually improve a fine-tuned policy model.
In parallel, DeepSeek-Prover V2 leverages a large language model to recursively decompose theorems into subgoals, combining this with reinforcement learning shaped by verifier feedback. Both approaches treat the Lean environment as an interactive medium and fully disregard natural language during inference.

TODO: transition from working in lean space to working in natural language space (using theoremllama paper)

Earlier efforts, such as TheoremLlama \cite{wang2024theoremllamatransforminggeneralpurposellms}, demonstrated that even mid-sized open models can reach strong formal proving performance when trained on bootstrapped Lean–natural language pairs.

\textbf{Integrating Graphs and forming proofs in natural language}
Our model extends prior work by integrating a ProofWiki-derived knowledge graph with large language models for automated proof generation. Using natural language as an intermediate representation allows access to a much broader corpus of LaTeX-based and informal proofs than formal codebases like Lean. It also harnesses LLMs’ emergent reasoning abilities and exposes interpretable reasoning traces that can reveal novel strategies. The tradeoff is added error and complexity in the informal-to-formal translation, especially in semantically precise edge cases.

We address this with a two-agent system for informal proof generation and formalization, supported by retrieval-augmented generation over graph-structured knowledge. This follows trends in autoformalization seen in DeepSeek-Prover-V1.5, which combines RL and tree search, and TheoremLlama, which shows gains from natural language intermediaries. Our graph-based retrieval also aligns with work like GraphRetriever and QA-GNN, where structure enables targeted, interpretable context. Iterative refinement and verification loops reflect recent advances in dynamic test-time compute \citep{Snell2024ScalingLLMTestTime}. Together, these elements advance scalable, interpretable, modular theorem proving with LLMs.

\section{Methodology}
\begin{figure*}[ht] \centering \includegraphics[width=\linewidth]{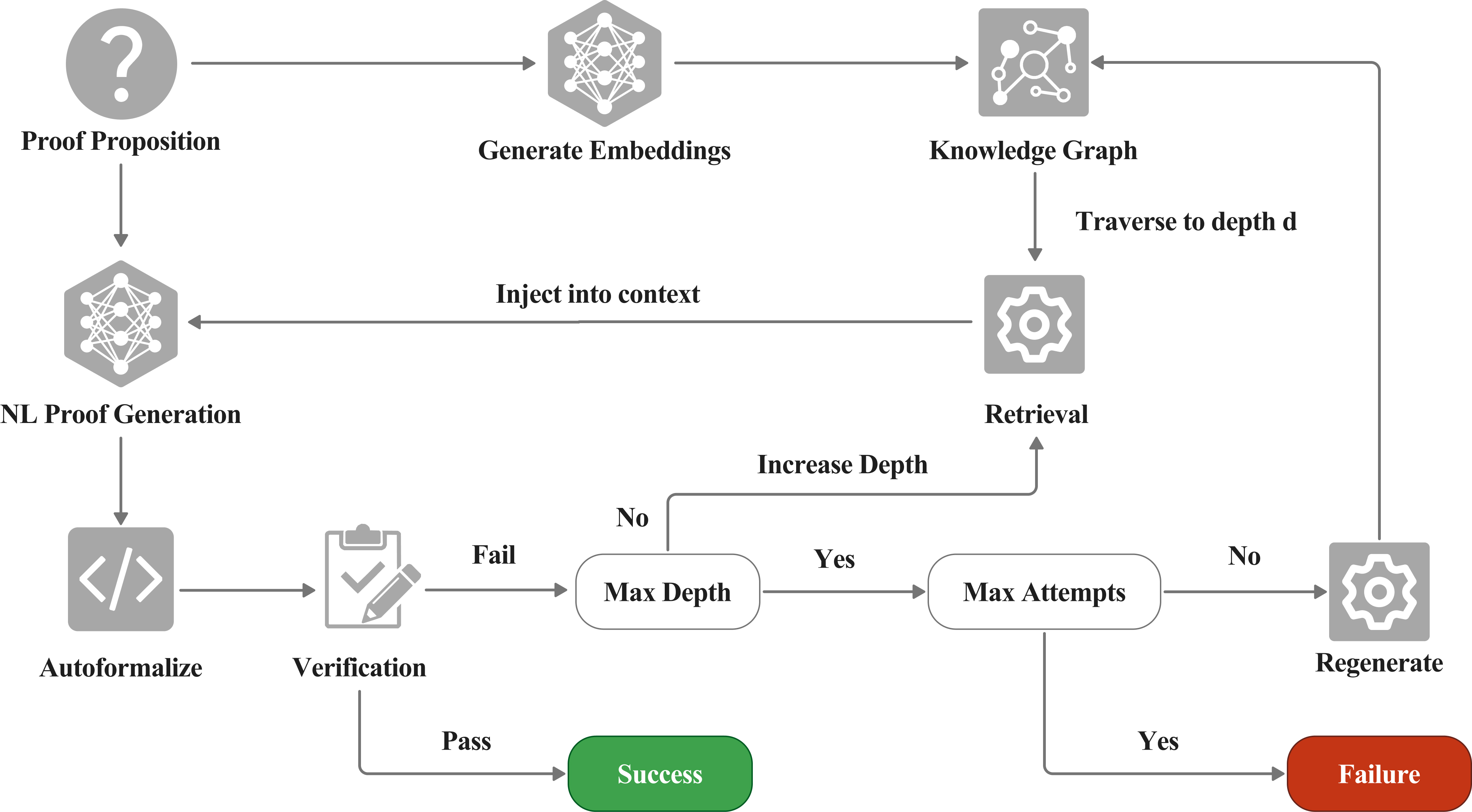} \caption{Whereas many modern proof systems focus on training time improvements, we integrate Node retrieval based on an interconnected knowledge graph into our proof system at inference time. Before generating a proof, we inject the most similar nodes into the context, then verify the proof using Lean. If the verification is unsuccessful, we grant the model the chance to traverse the graph deeper, where the knowledge graph allows it to explore other related concepts and theorems, on multiple attempts.} \label{fig:workflow} \end{figure*}

Our framework automates mathematical proof generation by integrating LLMs with a knowledge graph constructed from ProofWiki. We employ a multi-stage approach combining retrieval-augmented generation with a multi-step LLM system for proof formalization. The system consists of four main components: 1) knowledge graph traversal, 2) informal proof generation, 3) formal proof generation, 4) verification and refinement. Figure~\ref{fig:workflow} illustrates the overall KG-Prover workflow.




\subsection{Knowledge Graph Construction}
We built the underlying KG-Prover knowledge graph by parsing ProofWiki, an online compendium organized into distinct namespaces such as "Definition," "Axiom," and "Proof." By targeting these namespaces, we reliably extracted the formal components: the precise definition statements, axiom listings, theorem propositions (including lemmas and corollaries), and their corresponding proof details \citep{ProofWiki}. In our pipeline, each node represents a self-contained mathematical assertion--such as the text of a definition, the formal proposition of a theorem or lemma, or a corollary--while the textual proof that follows is stored as a property or linked entity. Hyperlinks within pages (e.g. references to earlier theorems or definitions) become edges in the graph, capturing higher level dependencies and conceptual relationships. We store this graph in a Neo4j \citep{Webber2012} graph database, augmenting each node with precomputed embedding vectors using OpenAI’s text-embedding-3-large model. An example entry from our nodes collection can be found in Appendix \ref{appendix:nodes_example}. This structure enables efficient semantic queries: given a problem statement, we compute its embedding and retrieve the top-$k$ most similar nodes to provide as a starting point for subsequent graph traversals. Explicit details on the construction of the knowledge graph can be found in Appendix \ref{appendix:dataset_sample}.

\subsection{KG-Prover}
Our KG-Prover framework consists of four main components: 1) knowledge graph traversal, 2) informal proof generation, 3) formal proof generation, 4) verification and refinement.

\subsubsection{Retrieval}

Let \( G = (V, E) \) be a knowledge graph, where \( V \) represents all nodes as mathematical theorems and \( E \) represents the edges between them. Given a proposition \( P \) that we are tasked to prove, we use the below-signified similarity function that assigns a relevance score to each node based on its similarity to \( P \).

We opt for cosine similarity by generating an embedding vector for \( P \), \( \mathbf{v}_P \), and comparing \( \mathbf{v}_P\) to the other node embeddings \( \mathbf{v}_i \in V \) in the knowledge graph:
    \begin{align*}
        S = sim(v_P,v_i)=\frac{v_P\cdot v_i}{\|v_P\|_2\|v_i\|_2}
    \end{align*}

If $P$ is not solved in the first iteration of generation, we introduce a depth parameter \( d \) that can be incremented up to an allowed depth \( D \). We iteratively expand the context by selecting up to $k$ additional nodes that are related concepts of previously selected nodes. \[k_1, k_2, \dots ,k_i = \arg\max_{V_{d-1} \in V} S(V_d, V_{d-1})\] where $V_{d-1}$ represents the set of all traversed nodes and $V_d$ represents the set of all 1-hop neighbors of $V_{d-1}$. Mathematically $V_{d-1}$ is defined as the set of $v_j:\exists (a, b) \in E \text{ where } a \in V_{d-1}$. For each traversal, we select the top k 1-hop neighbors ranked by our similarity function S.

This expansion continues until either:
\begin{itemize}
    \item \( P \) is resolved by the language model.
    \item The maximum depth \( D \) is reached and the amount of regenerating tries is expended.
\end{itemize}


\subsubsection{Informal Proof Generation}
Informal proof generation integrates retrieved-context into the language model prompt and uses the LLM to create an informal, natural language proof based on this enhanced input. Initially, generating a natural language proof allows KG-Prover to take advantage of the repository of known proofs outlined within the knowledge graph. If the proof fails to pass the verification stage, the framework iteratively deepens the context by one level in the knowledge graph, selecting the top-$k$ semantically closest neighboring nodes to uncover missing key concepts. The updated context is then used for subsequent proof generation attempts.

\subsubsection{Formal Proof Generation}

The autoformalization LLM ingests the code prefix, proposition, and informal proof before generating the translated formal proof. The framework for both the Autoformalizer and the generator models can be found in Appendix \ref{appendix:prompt_example}. Finally, the autoformalizers' output is parsed to extract well-specified Lean 4 code.

\subsubsection{Verification and Refinement}
To ensure the formal correctness of the proofs generated by our framework, we adopted the Lean verification method from DeepSeek-Prover-V1.5 to enhance the formalization step in our proof generation process, utilizing RLPAF to refine our model's ability to generate proofs that are verifiable in Lean \citep{Xin2024DeepSeekProverV15HP}. By integrating proof-assistant feedback, our models are more robust in producing proofs that adhere to the strict syntactic and logical requirements of Lean.

The formal proofs were verified using Lean 4 to ensure correctness. The generated proof code was submitted to Lean, and the results were analyzed. On failure, we extract error messages and feed them back into the autoformalizer along with adjusted prompts. This loop repeats until verification succeeds or we exhaust the allowed attempts $r$. If the maximum depth is not yet reached, we perform another graph traversal and generation loop.
\subsection{Scaling KG-Prover}
Self-consistency has proven itself as strongly effective, on commonly used reasoning as well as mathematical tasks, making use of the different approaches a language model might take while sampling multiple responses \citep{wang2023selfconsistencyimproveschainthought}. To utilize this phenomenon and improve robustness and accuracy under limited graph traversals, we incorporate test-time scaling, sampling, and search strategies:

\subsubsection{Best-of-N}
 For each proof task, we generate $n$ independent informal proofs, autoformalize, and verify them. A dedicated model then acts as a judge, evaluating each candidate's proof across dimensions of mathematical correctness, clarity, and reasoning completeness. The judge assigns scores from 0-10 and provides justification for each evaluation. Candidates are then sorted by their scores, with the highest-scoring proof selected as the "optimal" solution to convert into Lean.

\subsubsection{Beam Search}
We organize proof candidates into a beam of width $w$ and search depth $s$. At each step, the top-$w$ candidates are expanded by generating refinements based on verifier feedback \citep{sun-etal-2023-allies}. These refinements are then scored and ranked, with the top-$k$ candidates retained for subsequent iterations. The process repeats $s$ times, ultimately returning the "best" proof that is both high-quality in terms of interpretability and formally verifiable. This balances the exploration of diverse proof paths with verification-driven refinement.

\section{Experiment Design}
\subsection{Models}
To create semantic representations in the form of embeddings, we used OpenAI's \texttt{text-embedding-3-large} model \citep{Neelakantan2022}.

For informal proof generation, we utilized GPT-4o-mini, as well as Claude 3.5 Sonnet and a collection of LLAMA 3 models \citep{OpenAI2024, anthropic2024claude35sonnetaddendum, grattafiori2024llama3herdmodels}. We measure performance on the COT-reasoning models Deepseek-R1, o1-mini, and o4-mini \citep{deepseekai2025deepseekr1}.

As an Autoformalizer we use DeepSeek-Prover-V1.5 \citep{Xin2024DeepSeekProverV15HP}, which is an open-source language model, designed for theorem proving in Lean \citep{Moura2021}. We use the model explicitly only for the translation of the already generated informal proof into Lean format to validate informal proofs.

\subsection{Datasets}
To evaluate the effectiveness of our framework, we conducted experiments on multiple benchmarks commonly used in automated theorem proving: \textbf{miniF2F}\footnote{Unless stated otherwise, references to miniF2F denote the average performance across only the test split.}, \textbf{ProofNet}, and \textbf{MUSTARDSAUCE} \citep{Zheng2022, Azerbayev2023,huang2024mustardmasteringuniformsynthesis}.  MiniF2F is a benchmark dataset of formal mathematics problems sourced from undergraduate-level mathematics competitions. ProofNet is a large-scale dataset of mathematical proofs and theorem statements, ranging in difficulty and domain. MUSTARDSAUCE is the dataset MUSTARD generated itself using GPT-4.

Our exact dataset configuration can be found in Appendix \ref{appendix:benchmarks}.

\subsection{Introduced Scaling Parameters}
For our evaluations, we introduce multiple parameters that can be varied. In our evaluations:
\begin{itemize}
    \item $k$ signifies the number of selected nodes from the current depth descending based on semantic similarity.
    \item $r$ signifies the provided number of attempts on one individual proof.
    \item $d$ defines the depth the retriever is allowed to traverse in the knowledge graph.
    \item $n$ defines the number of candidates generated by best of N
    \item $w$ or $beam$ defines the width of the beam for the beam search implementation
    \item $search\_depth$ defines the depth during beam search
\end{itemize}

\begin{table*}[h!]
\centering
\begin{tabular}{l l|c c c c c c}
\toprule
\textbf{Dataset ($\uparrow$)} & \textbf{Method} & \makecell{\textbf{Claude 3.5}\\\textbf{Sonnet}} &
\makecell{\textbf{Deepseek}\\\textbf{R1}} &
\makecell{\textbf{Llama 3.1}\\\textbf{8B}} &
\makecell{\textbf{Llama 3.3}\\\textbf{70B}} &
\textbf{GPT 4o} &
\makecell{\textbf{o1}\\\textbf{-mini}} \\
\midrule
\multirow{3}{*}{\textbf{ProofNet}}
 & Base   & 2.69\% & 2.69\% & 3.76\% & 2.15\% & 3.23\% & 3.76\% \\
 & RAG    & 3.76\% & 3.76\%& 3.76\% & 3.76\%  & 5.38\% & 5.91\% \\
 & KG-Prover & 4.84\% & 5.38\%& 4.30\%  & 4.30\%  & 6.45\% & \textbf{6.99\%} \\
\midrule
\multirow{3}{*}{\textbf{miniF2F}}
 & Base   & 22.95\% & 20.08\% & 20.49\% & 25.00\% & 23.36\% & 23.77\% \\
 & RAG    & 28.69\% & 22.54\% & 24.59\% & 24.59\%  & 28.69\% & 28.28\% \\
 & KG-Prover & 31.15\% & 28.28\%& \textbf{31.97\%}  & 30.74\%  & 30.74\% & 30.74\% \\
\midrule
\multirow{3}{*}{\textbf{MUSTARD}}
 & Base   & 28.00\%& 20.00\%& 24.00\%& 25.60\%& 28.00\%& 24.80\%\\
 & RAG    & 28.40\%& 25.00\%& 28.00\%& 28.8\%& 28.00\%& 26.80\%\\
 & KG-Prover & 30.00\%& 27.00\%& 27.60\%& 32.5\%& 30.00\%& \textbf{34.00\%}\\
\bottomrule
\end{tabular}
\caption{Comparison of models across ProofNet, miniF2F, and MUSTARDSAUCE datasets. Accuracy scores reflect the performance of a single run with a maximum of three attempts per proof, measured as a percentage of successful proof generations. The bolded numbers show the largest performance gain from baseline to our KG-Prover for each dataset, achieving more than 11\% gain.}
\label{tab:model_dataset_accuracy}
\end{table*}

\section{Results}

\subsection{Knowledge Graph Performance}
As visualized in Table \ref{tab:model_dataset_accuracy}, using graphs consistently outperforms baseline proof systems and over Retrieval Augmented Generation. Performance gains of the KG-Prover ranged from 2-11\% across different models\footnote{Although $top-k=5$ is a fixed parameter, the actual value can be smaller depending on the number of related nodes available at the current depth.}. Llama 3.1 8B achieved a 31.97\% success rate on miniF2F, compared to a 20.49\% baseline.

ProofNet represents the most challenging dataset with the lowest overall performance (2-7\% success rates), attributed to the difficulty of the problems. They require higher abstract mathematical reasoning and more intricate proof structures. The miniF2F dataset showed moderate performance (20-31\% success) because it includes more structured mathematical problems, intermediate complexity of proofs, and more predictable reasoning patterns.

MUSTARDSAUCE demonstrated moderate performance as well (24-34\% success).




\subsection{Finetuned foundation models}

\begin{table}[H]
\centering
\begin{tabular}{l c}
\toprule
\textbf{Model} & \textbf{minif2f} \\
\midrule
TheoremLlama (pass@128) & 35.04\% \\
TheoremLlama + KG-Prover & \textbf{36.89\%} \\
\bottomrule
\end{tabular}
\caption{Using finetuned models with knowledge graph traversal depth = 6, witnesses improved performance over 128 rounds of generation}
\label{tab:pass128-model-comparison}
\end{table}

As visualized in Table \ref{tab:pass128-model-comparison}, using structured Knowledge of our KG-Prover with a depth of 6, performs 1.85 percentage points better than a finetuned model outperforming the pass@128\footnote{We follow the definition of pass@k defined by \citet{chen2021evaluatinglargelanguagemodels}} on finetuned Lean provers.

\subsection{Scaling Traversal Depth}
To allow the model for failure correction and improvement, the graph system has multiple consecutive attempts defined as $r$. Each attempt allows the model to traverse further in the graph and explore more nodes.

\begin{figure}[htb]
  \centering
  \begin{tikzpicture}
    \begin{groupplot}[
        group style={%
          group size=2 by 2,
          horizontal sep=15pt,
          vertical sep=15pt,
        },
        width=0.58\columnwidth,      
        height=0.55\columnwidth,
        tick label style={font=\scriptsize},
        label style={font=\normalsize},
        title style={font=\normalsize},
        every axis plot/.append style={thick},
        xmin=1, xmax=7,
        xtick={1,...,7},
        ylabel={Accuracy (\%)},
        cycle list={
          {blue,mark=o},
          {orange,mark=triangle*},
          {blue,dashed,mark=none},
          {orange,dashed,mark=none}
        }
    ]

    \nextgroupplot[
      title={minif2f},
      ymin=20, ymax=36,
    ]
      \addplot coordinates {(1,24.59) (2,29.10) (3,31.56) (4,32.38) (5,33.61) (6,34.84) (7,35.25)};
      \addplot coordinates {(1,25.82) (2,30.33) (3,32.38) (4,33.28) (5,34.02) (6,34.02) (7,34.84)};
      \addplot coordinates {(1,20.49) (7,20.49)};
      \addplot coordinates {(1,23.77) (7,23.77)};

    \nextgroupplot[
      title={ProofNet},
      ymin=2.5, ymax=9,
      ylabel={} 
    ]
      \addplot coordinates {(1,2.96) (2,3.76) (3,4.30) (4,4.30) (5,4.84) (6,4.84) (7,5.38)};
      \addplot coordinates {(1,4.30) (2,5.91) (3,6.45) (4,6.99) (5,6.99) (6,7.53) (7,8.06)};
      \addplot coordinates {(1,3.76) (7,3.76)};
      \addplot coordinates {(1,3.76) (7,3.76)};

    \nextgroupplot[
      title={MUSTARDSAUCE},
      ymin=14, ymax=43,
  title style={at={(axis description cs:0.5,-0.25)},anchor=north},
      ylabel={}
    ]
      \addplot coordinates {(1,14.40) (2,26.40) (3,30.40) (4,33.60) (5,35.60) (6,37.20) (7,38.00)};
      \addplot coordinates {(1,18.00) (2,26.80) (3,32.40) (4,36.80) (5,38.40) (6,40.40) (7,41.60)};
      \addplot coordinates {(1,24.00) (7,24.00)};
      \addplot coordinates {(1,24.80) (7,24.80)};

    \nextgroupplot[
      hide axis,
      xmin=0, xmax=1, ymin=0, ymax=1,
      legend style={
        font=\small,
        at={(0.5,0.4)}, anchor=center,
        legend columns=1
      }
    ]
      \addlegendimage{blue,mark=o}
      \addlegendentry{LLAMA}
      \addlegendimage{orange,mark=triangle*}
      \addlegendentry{o1-mini}
      \addlegendimage{blue,dashed}
      \addlegendentry{LLAMA Base}
      \addlegendimage{orange,dashed}
      \addlegendentry{o1-mini Base}

    \end{groupplot}
  \end{tikzpicture}
  \caption{Accuracy increases with greater traversal depth $r$ in the knowledge graph}
  \label{fig:depth-results}
\end{figure}

As more proofs get injected into the context and the model gets more tries to correct initial mistakes, the accuracy scales higher per iterative refinement step. This effect is most predominant in smaller parameter models, such as Llama 3.1 8b. This behavior is captured in Figure \ref{fig:depth-results}. We can see that with more nodes injected, the performance rises.

\begin{figure*}[htb]
  \centering
  \begin{tikzpicture}
    \begin{groupplot}[
        group style={
          group size=3 by 1,
          horizontal sep=1.0cm,
        },
        width=0.31\textwidth,
        height=0.29\textwidth,
        tick label style={font=\scriptsize},
        label style={font=\normalsize},
        title style={font=\normalsize},
        every axis plot/.append style={thick},
        xmin=1, xmax=7,
        xtick={1,...,7},
        xlabel={Traversal Depth $r$},
        ylabel={Accuracy (\%)},
        legend style={
          font=\small,
          at={(0.5,-0.35)},
          anchor=north,
          legend columns=2,
        },
        cycle list={
          {blue,mark=o,mark options={scale=1.2}},
          {orange,mark=triangle*,mark options={scale=1.2}}
        }
    ]

    \nextgroupplot[
      title={miniF2F},
      ymin=30, ymax=55,
    ]
    \addplot coordinates {
      (1,31.15) (2,36.48) (3,38.52) (4,39.34)
      (5,40.57) (6,40.98) (7,41.34)
    };
    \addplot coordinates {
      (1,39.34) (2,43.85) (3,46.72) (4,48.77)
      (5,49.59) (6,51.64) (7,52.87)
    };

    \nextgroupplot[
      title={ProofNet},
      ymin=5, ymax=15,
      ylabel={},
    ]
    \addplot coordinates {
      (1,5.38) (2,7.53) (3,8.60) (4,8.60)
      (5,9.14) (6,9.68) (7,10.75)
    };
    \addplot coordinates {
      (1,6.99) (2,9.68) (3,10.75) (4,11.29)
      (5,11.29) (6,12.37) (7,13.44)
    };

    \nextgroupplot[
      title={MUSTARDSAUCE},
      ymin=25, ymax=65,
      ylabel={},
    ]
    \addplot coordinates {
      (1,32.40) (2,40.80) (3,44.80) (4,45.60)
      (5,47.20) (6,49.60) (7,50.40)
    };
    \addplot coordinates {
      (1,29.60) (2,40.00) (3,46.80) (4,51.20)
      (5,54.00) (6,57.60) (7,60.00)
    };

    \legend{LLaMA 8B, o4-mini}
    \end{groupplot}
  \end{tikzpicture}
  \caption{Comparing different depths for the beam search method on a set of parameters that are n = 5, beam width 3, search depth 2.}
  \label{fig:all_together}
\end{figure*}
\subsection{Combined Scaling with Beam Search}



As visualized in Figure \ref{fig:all_together}, combining the knowledge graph with approaches that can sample responses from the context of the KG-Prover has a positive effect on accuracy across all benchmarks. Achieving up to 10.75\% on ProofNet, 41.34\% on miniF2F and 50.40\% on MUSTARDSAUCE, which equates to improvements of 26.40\% in the highest scaling configuration.
Across all three dataset we find the first three depth increases to be the most effective in scaling the accuracy. While the leaps in accuracy flatten towards deeper depths, on harder datasets, we see the higher depths actually still bring a consistent improvement.

Additionally, we see that even on depth one the performance consistently beats the baseline and on average performs on  par\footnote{considering slight deviations of $\pm$ 1\%} better than the KG-Prover without multiple candidate proofs.

\subsection{Performance Tradeoffs}
We isolate the cost that dominates monetary expenditure in practice: the
number of language-model API calls and the prompt/response tokens for those calls
consume. Autoformalization and Lean verification are performed on local
hardware and are therefore ignored in this section.

\paragraph{Notation}
Let $T_{\mathrm{p}}$ be the average prompt length (in tokens) for a single
informal-proof request and $T_{\mathrm{r}}$ the average length of the LLM’s
response.
Our methods differ only in how many \emph{times} that request–response pair
is issued.

\begin{table*}[h]
\centering
\small
\setlength{\tabcolsep}{6pt}
\begin{tabular}{lccc}
\toprule
\textbf{Setting} &
\textbf{\#\,completions} &
\textbf{Token budget} &
\textbf{Cost multiplier} \\
\midrule
\textsc{Base} &
$1$ &
$T_{\mathrm{p}}+T_{\mathrm{r}}$ &
$1\times$ \\
\textsc{KG-Prover} &
$r$ &
$r\,(T_{\mathrm{p}}+T_{\mathrm{r}})$ &
$r\times$ \\
\textsc{KG-Prover + Beam} &
$r\!\Bigl(n+w\sum\limits_{i=0}^{s-1}w^{\,i}\Bigr)$ &
$\displaystyle
r\Bigl(n+w\frac{w^{s}-1}{w-1}\Bigr)\!(T_{\mathrm{p}}+T_{\mathrm{r}})$ &
$\displaystyle
r\Bigl(n+w\frac{w^{s}-1}{w-1}\Bigr)\!\times$ \\
\bottomrule
\end{tabular}
\caption{Language-model usage per problem instance
($r$ traversal attempts, initial beam of $n$ candidates, beam width $w$, tree depth $s$).}
\label{tab:llm-calls}
\end{table*}
\paragraph{Implications}
\begin{itemize}\setlength{\itemsep}{2pt}
\item \textbf{Retries scale linearly.}  Each additional attempt multiplies
  total spend by $r$ but yields diminishing accuracy gains beyond
  $r{=}3$ (\autoref{fig:depth-results}).
\item \textbf{Beam search is the price driver.}
  The geometric factor
  $n+w(w^{s}-1)/(w-1)$ explodes quickly:
  doubling the beam width from $3$ to $6$ would more than triple token
  usage while delivering $\le$1 pp extra accuracy on \textsc{ProofNet}.
\end{itemize}

\section{Conclusion \& Discussion}
We present a framework that automates mathematical proof generation by integrating LLMs with a knowledge graph to utilize inter-dependencies across mathematical proofs. Our approach demonstrates the potential of combining multiple mathematical concepts in an intertwined graph. By doing so, language models can be effectively guided toward correct proof generation, resulting in improved accuracy and enhanced abilities in both reasoning through and formalizing proofs in natural language, whilst adding lean verification in a separate translation step.

We establish that existing foundation models can achieve similar or higher performing results as fine-tuned models, by simple context injections of related concepts during inference time, without requiring any additional pre-training, expert iteration, or training system of any kind. By doing this we witness performance increases across datasets of up to 11\% by just using the KG-Prover and up to 26\% when combined with proper scaling techniques.

\section{Limitations}
Despite the advancements in capturing semantic relationships in text via vectorized embeddings, embeddings can potentially suffer from issues such as loss of fine-grained logical structure and difficulties in preserving contextual dependencies across larger passages. This can lead to challenges in accurately retrieving relevant mathematical statements, especially in formalized settings where precise definitions and logical consistency are crucial. While we filter and discard irrelevant details, signs, and other minutiae, XML dumps can introduce noise that might disrupt of affect the semantic search and embeddings.

While our approach successfully formalizes proofs from structured datasets, its performance on entirely novel or highly abstract mathematical problems remains uncertain. Models trained on existing proofs may struggle with creative problem-solving or unconventional mathematical approaches.

Large Language Models have finite context windows, meaning lengthy or complex proofs may exceed the model's processing capacity. This might result in incomplete reasoning, loss of critical details, or forgetting earlier steps in multi-stage proofs.

Future work may enhance the knowledge graph and improve the autoformalization process to handle more complex mathematical concepts.

\section{Reproducibility Statement}
Our experiments were conducted using publicly available Datasets and Models. GPT-4o, 4o-mini, o1-mini and text-embedding-3-large can be accessed via \url{https://openai.com/api/}. Both Deepseek-R1 and the LLAMA 3 collection are open-sourced models. Claude models can be accessed via their respective API endpoints, under \url{https://www.anthropic.com/api}.

\hyperlink{https://github.com/zhangir-azerbayev/ProofNet}{ProofNet} and \hyperlink{https://github.com/openai/miniF2F}{miniF2F}, and \hyperlink{https://github.com/Eleanor-H/MUSTARD}{MUSTARDSAUCE} are publicly available datasets. Our Code is publicly available on GitHub; we encourage anyone to validate and extend our findings.
The Neo4j-based graph database can be used under \url{https://neo4j.com} and could potentially be replaced with alternative graph databases as desired.

\section{Ethical Considerations \& Risks}
Our knowledge base is derived from ProofWiki, an open database for formal proofs. While the page is moderated, adversaries could attempt to incorporate harmful content or incorrect factual information into the extracted pages. However, we consider this risk to be unlikely.

Although alignment work continues to progress Large Language Models can introduce biases towards certain marginalized groups or other minorities. All of our introduced models are moderated and have content filters that should prevent models from generating harmful content. However said filters aren't perfect, models can still be exploited via sophisticated prompting and other adversarial techniques. Given our contribution to the framework, we expect no increased risk in any of the given safety evaluation measures proposed.

\subsection{GPU usage}

\begin{table}[h]
    \centering
    \begin{tabular}{c|c|c}
        GPU model & Watts & approx. usage Time \\
        \midrule
        Nvidia A40 & 300 W & 700 hours \\
        Nvidia RTX A5000 & 300 W & 50 hours \\
    \end{tabular}
    \caption{Estimated GPU usage for all Evaluations.}
    \label{tab:my_label}
\end{table}

The shown GPU usage may only partially reflect an accurate measure of the computational resources required, as major models are only available through API endpoints. We estimate the inference time on said APIs to be roughly 170 hours.

\bibliographystyle{plainnat}
\bibliography{custom}

\clearpage

\appendix

\section{Using Lean provers for informal proof generation}
As discussed, our approach utilizes a two step process that generates an informal proof in natural language that is then translated and validated in Lean. As visualized in Table \ref{tab:wrapped_model_comparison}, even Lean provers can be enhanced using the KG-Prover that utilizes natural language.

\begin{table}[h!]
\centering
\begin{tabular}{lcc}
\toprule
\textbf{Method} &
\makecell[c]{\textbf{Theorem}\\\textbf{Llama}} &
\makecell[c]{\textbf{DeepSeek}\\\textbf{Prover-V1.5}} \\
\midrule
minif2f Base & 32.38 & 35.75 \\
minif2f RAG & 34.84 & 36.48 \\
minif2f KG-Prover & \textbf{36.89} & \textbf{37.71} \\
\bottomrule
\end{tabular}
\caption{Lean-based provers show increased performance using the KG-Prover (traversal depth = 6) even when the Node data is in natural language.}
\label{tab:wrapped_model_comparison}
\end{table}

This phenomenon demonstrates that even Lean-optimized and fine-tuned systems benefit from structured natural language knowledge encompassing related proofs.

\section{Structural Improvement}
Few-shot learning, even with briefly related examples, has been shown to improve performance across a variety of tasks and domains.

Therefore we hypothesize that even only partly related proof nodes will improve not only the proof understanding but will also benefit the structured formalization that is required for the correct interpretation and conversion of informal natural language into Lean4.

\section{Judging the Best of $N$ Tries}
Interestingly, the results in Table \ref{tab:best_of_n_base} reveal a non-linear relationship in more challenging datasets like ProofNet, where an intermediate value (e.g., $N=6$) did not always outperform a lower or higher $N$. This suggests that simply increasing the number of candidates is not universally beneficial; the quality of each candidate and the effectiveness of the judging mechanism play critical roles. As such, finding the right balance in model temperatures is crucial because an optimal setting enhances the judging process by providing a diverse pool of high-quality candidates\footnote{Both best of $N$ and best of $N$ + tree search method evaluations had LLama 3.1 8B set on a temperature of 0.7}.

\begin{table}[ht]
\centering
\begin{tabular}{l|l|ccc}
\toprule
\multirow{2}{*}{\textbf{Dataset}} & \multirow{2}{*}{\textbf{Model}}
  & \multicolumn{3}{c}{\textbf{Best of N}} \\
\cmidrule(lr){3-5}
 & & N=2 & N=6 & N=10 \\
\midrule
ProofNet
  & Llama 8B & $6.45\%$ & $5.38\%$ & $8.60\%$ \\
\midrule
miniF2F
  & Llama 8B & $30.33\%$ & $30.74\%$ & $31.97\%$ \\
\midrule
Mustard
  & Llama 8B & $30.00\%$ & $32.80\%$ & $33.6\%$ \\
\bottomrule
\end{tabular}
\caption{Results by dataset \textit{with the graph approach}, comparing “Best of $N$” values between 2 and 10.}
\label{tab:best_of_n_base}
\end{table}



\section{Deterministic Evaluations}
Unless specified otherwise, we use greedy decoding for all of our experiments. Additionally, the semantic search in our Graph knowledge base will yield identical outputs, given that the input doesn't change between different runs.

While this behavior can be favorable in some situations, other evaluations may benefit from slight variations in different seeds. To introduce a slight stochasticity, other evaluations may vary the temperature parameter of the employed models, and use the introduced method in Appendix \ref{appendix:knowledgeGraphStochastic} to introduce randomness into our knowledge graph.
\subsection{Knowledge Graph Stochasticity}
\label{appendix:knowledgeGraphStochastic}
To mitigate fully repetitive outputs Nodes from the knowledge graph, we propose top-k shuffling, where we retrieve the k-highest ranked nodes, shuffle them, and select a subset. This method ensures diversity in individual generations. We favor this implementation over random sampling over a broader set of candidate nodes, selecting from a pool beyond the strict top-k. Due to the potentially less relevant knowledge, trading off precision for increased coverage.

The level of stochasticity can be tuned dynamically based on confidence scores or response variance metrics

\section{Examples}
\subsection{Node example}
\label{appendix:nodes_example}

\begin{itemize}
    \item \textbf{from\_id}: The ID of the current node.
    \item \textbf{to\_id}: The ID of the linked node (found using the title-name-to-ID mapping).
    \item \textbf{type}: There are 6 different relationship categories:
    \end{itemize}
    \begin{verbatim}
    USES_DEFINITION,
    RELATED_DEFINITION,
    USES_AXIOM,
    SIMILAR_PROOF,
    PROOF_DEPENDENCY,
    PROOF_TECHNIQUE.
    \end{verbatim}

\subsection{Prompt Examples}
\subsubsection{Prompt Example 1}

 The model was provided with the informal proof and a code template, and it generated the corresponding formal proof in Lean 4. Each element was processed to extract the title, namespace, and content.

\label{appendix:prompt_example}

\begin{verbatim}
You are a Lean 4 code generator.
We have:
HEADER:
{header}

INFORMAL PROOF:
{informal_proof}

PREFIX:
{informal_prefix}

STATEMENT:
{formal_statement}

GOAL (optional):
{goal}

INSTRUCTIONS:
1. Output exactly one triple-backtick code
block containing valid Lean 4 code.
2. Do not include any text or explanations
outside the code block.
3. Make sure it compiles in Lean 4.

Required Format:
# Start
```lean4
<Lean code here>
```
# End
\end{verbatim}

\subsubsection{Prompt Example 2}
\label{appendix:prompt_example2}
\begin{verbatim}
You are a mathematics expert focused on
generating clear informal proofs.

Given the following mathematical problem
and context, generate a clear and detailed
informal proof in natural language.

Context: [Retrieved context]

Problem: [Problem statement]

Provide your proof in the following format:

Informal Proof:
[Your proof here] \end{verbatim}

\section{Graph Dataset}
\label{appendix:dataset_sample}

We parsed an XML dump of ProofWiki, where each \texttt{<page>} element was processed. Irrelevant sections were filtered, and the wikitext was cleaned to obtain structured content.
\subsection{Node structure}
We represented each mathematical concept as a node in the knowledge graph, storing attributes such as:

\begin{itemize}
    \item \textbf{id}: Unique identifier.
    \item \textbf{type}: Content type (e.g., definition, theorem).
    \item \textbf{title}: Page title.
    \item \textbf{name}: Extracted from the title.
    \item \textbf{content}: Theorems in algebraic notation.
\end{itemize}

\section{Benchmarks}
All datasets present their samples with natural language and a formal statement in Lean, which we use as ground truth to compare against.

\label{appendix:benchmarks}
By utilizing miniF2F, ProofNet, and MUSTARDSAUCE, we assess our framework's ability to generate and formalize proofs across diverse mathematical problems. The datasets provided a standardized evaluation setting, allowing us to compare our results uniformly with existing approaches and to analyze the strengths and limitations of our Method. However, it is possible that our setup deviates from the ones introduced in the respective papers of the dataset, which explains the varied performance across tasks, which is especially apparent on MUSTARDSAUCE. To set up a comparable evaluation, we compute the baseline of our setup as well, rather than taking the previous State-of-the-Art.
\subsection{Used splits}
We ran 186 problems from the test split of ProofNet, 244 problems from the test split of miniF2F, and randomly selected 250 theorem-proving problems from MUSTARDSAUCE.

\section{Search Strategies within the Knowledge Graph}

To optimize the process of automated proof generation, we explored different methods for navigating the constructed knowledge graph. Specifically, we implemented two primary search strategies: Breadth-First Search (BFS) and semantic search using vector embeddings. This section elaborates on these methodologies, their implementation in our framework, and analyzes their respective advantages and disadvantages in our scenario.

\subsection{Breadth-First Search (BFS)}

Breadth-First Search is a classic graph traversal algorithm that systematically explores the vertices of a graph in layers, starting from a given root node and expanding outward to neighboring nodes at increasing depths.
In our framework, BFS was utilized as follows:

\begin{enumerate}
    \item \textbf{Zero-Shot Prompting}: We initially present the problem statement directly to the GPT model without any additional context, requesting a proof in a zero-shot setting.
    \item \textbf{First-Level Traversal}: If the zero-shot attempt is unsuccessful, we perform a BFS to explore the immediate neighboring nodes of the problem statement node. Specifically, we retrieve up to the nearest 50 nodes connected directly to the root node.
    \item \textbf{Contextual Prompting}: We then prompt the GPT model again, providing the problem statement along with the content from the retrieved neighboring nodes to supply additional context for proof generation.
    \item \textbf{Iterative Expansion}: If the proof remains incomplete or incorrect, we extend the BFS to the next level by including nodes that are two edges away from the root, effectively expanding the context window before re-prompting the GPT model.
\end{enumerate}

The advantage of BFS is that it allows for a systematic exploration of the knowledge graph, ensuring that all nodes within a certain depth are considered, which may uncover relevant but non-obvious connections. By incrementally increasing the depth of traversal, we can control the amount of additional information provided to the GPT model, potentially improving the quality of the generated proof.

However, BFS can be computationally expensive, especially in densely connected graphs, as the number of nodes grows exponentially with each additional level of depth. Including a broad set of neighboring nodes may introduce irrelevant or redundant information, which could overwhelm the GPT model and hinder its ability to generate a coherent proof.

\subsection{Semantic Search Using Embeddings}

Semantic search leverages vector embeddings to identify nodes that are semantically similar to a given query \cite{Neelakantan2022}. Each node in our knowledge graph is associated with a high-dimensional embedding vector, enabling similarity computations.

 \textbf{Hierarchical Prompting}: Similar to the BFS approach, we begin with a zero-shot prompt. If unsuccessful, we incrementally include the most similar nodes into the context when re-prompting the GPT model, effectively performing one-shot, two-shot prompting, and so on.

Semantic search is computationally less intensive than BFS, as it avoids exhaustive traversal and focuses only on nodes with high semantic relevance. By prioritizing nodes that are semantically similar to the problem statement, we provide the GPT model with highly pertinent information, potentially improving proof generation quality. The disadvantages are that the effectiveness of semantic search is contingent upon the embedding model's ability to accurately capture mathematical semantics, which may be challenging for complex or abstract concepts. Important nodes that are not semantically similar based on the embedding (e.g., foundational axioms or lemmas) may be overlooked, potentially omitting crucial information required for the proof.

Regardless of the search method used, we adopted an iterative prompting strategy with the GPT model. This approach allows us to manage the amount of information provided to the GPT model, aiming to strike a balance between context richness and the model's capacity to process and utilize the information effectively.

\section{Failure Scenarios}

Although we see strong performance across multiple proof benchmarks, there are certain scenarios in which models \& techniques fail to function optimally. Across multiple runs, we found the following possible errors:
\begin{itemize}
    \item The informal proof is correct, but the conversion into a formal proof fails.
    \item The required knowledge is not in the graph, and other topics are too briefly related to be extrapolated.
    \
\end{itemize}

In our manual analysis, we found that approximately 35\% of the failures occur when the formal proof is incorrect despite the informal proof being largely valid. This suggests that the challenge often lies not in the mathematical reasoning itself, but in bridging the gap between informal and formal representations. Informal proofs frequently rely on high-level abstractions, implicit assumptions, or natural language shortcuts (e.g., “it follows that,” “by symmetry”) that do not always translate cleanly into Lean 4, which demands precision and fully explicit logic. Typical issues include omitted hypotheses, ambiguous theorem references, or imperfect formalization of induction and algebraic steps.

It is rare that traversal doesn't gather relevant information or that the knowledge is not available and only apparent on particularly hard questions.
However, for difficult questions, such as those proposed by the International Math Olympiad, the graph cannot find the most relevant nodes.

\end{document}